\documentclass[runningheads]{llncs}

\usepackage[T1]{fontenc}
\usepackage{graphicx}
\usepackage{color}
\usepackage{amsmath}
\usepackage{amssymb}
\usepackage{esvect}
\usepackage{algorithm}
\usepackage{algpseudocode}
\usepackage{booktabs}
\usepackage{array}
\usepackage{subcaption}
\newcolumntype{C}[1]{>{\centering\arraybackslash}p{#1}}

\begin{document}

\title{Cost-aware Stopping and Benchmarking for Evolutionary Optimization}



\author{
Xin Li\orcidID{0009-0003-6509-6197} \and
Juergen Branke\orcidID{0000-0002-4343-5878} \and
Xuan Vinh Doan\orcidID{0000-0002-8572-5208}
}

\authorrunning{X. Li et al.}

\institute{
Warwick Business School, University of Warwick,
CV4 7AL, Coventry, UK \\
\email{Xin.Li.2@warwick.ac.uk,
Juergen.Branke@wbs.ac.uk,
Xuan.Doan@wbs.ac.uk}
}
\maketitle

\begin{abstract}
In particular when function evaluations are costly, stopping an evolutionary algorithm at the right time can avoid unnecessary computational effort. This paper proposes a cost-aware stopping criterion for both single-objective EAs (SOEAs) and multi-objective EAs (MOEAs). The proposed rule predicts the improvement of further search and terminates when the expected improvement no longer justifies the cost of additional function evaluations.  We conduct experiments on both SOEA and MOEA settings and compare the proposed rule with existing stopping criteria using a metric that jointly accounts for solution quality and evaluation cost. The results show that the proposed stopping criterion achieves a better trade-off between solution quality and function evaluation cost, with performance close to the hindsight-best benchmark which has access to complete information.

\end{abstract}

\section{Introduction} \label{sec:intro}

An important practical question when applying an EA is when to stop. Unlike many classical optimization methods, EAs do not generally provide a natural certificate indicating that further computation will no longer improve the solution. Every EA has at least one stopping criterion \cite{liu2018termination}. A simple and widely used approach is to terminate the search after a predefined number of generations or function evaluations \cite{ghoreishi2017termination}. However, an insufficient budget may terminate the search before high-quality solutions are obtained, whereas an excessive budget may spend substantial computational resources for little additional improvement. Consequently, termination can be viewed as a trade-off between solution quality and the computational cost required to obtain further improvements \cite{liu2018termination}.

Liu et al.~\cite{liu2018termination} distinguish between a \emph{progress indicator}, which summarizes the current state or progress of the search, and a \emph{termination criterion}, which translates the observed progress into a stop-or-continue decision. Progress can, for example, be characterized through convergence, population diversity, solution quality, or changes between successive populations. Based on such information, termination decisions can be made using predefined thresholds, statistical inference, fuzzy methods, or other heuristic mechanisms. Ghoreishi et al.~\cite{ghoreishi2017termination} survey a wide variety of termination criteria, reflecting different notions of convergence and search progress. 

Although these approaches use increasingly sophisticated information about the optimization trajectory, many stopping rules are fundamentally designed to identify convergence or stagnation. 
However, detecting that progress has become small is not necessarily equivalent to further search not not being worthwhile. Even a small potential improvement may justify another generation when evaluations are inexpensive, whereas even a larger improvement may not be worthwhile when function evaluations are very costly. This motivates a stopping perspective that explicitly evaluates the expected benefit of continuing relative to its computational cost. 

In a closely related work, Kitamura and Tanabe benchmark existing stopping criteria for evolutionary multi-objective optimization by jointly considering solution quality and computational budget \cite{kitamura2026benchmarking}. Their study highlights the trade-off between stopping too early and wasting evaluations by stopping too late. However, they evaluate stopping criteria against a retrospectively identified best stopping point, defined as the last generation at which the best-so-far hypervolume improves along the observed search trajectory. 

Rather than asking only whether an EA has converged or whether its recent progress is sufficiently small, we ask whether \emph{the predicted improvement from one additional generation is worth its evaluation cost.} To answer this question, we develop a cost-aware stopping framework that uses the historical trajectory of the best-so-far solution(s). At each generation, a learning-curve model is fitted to the historical progress and used to predict the outcome of the next generation. The predicted one-generation improvement is then compared with the cost of performing the additional function evaluations. The search terminates when the predicted improvement can no longer justify this cost.

The same underlying principle is applied to both single- and multi-objective EAs. For SOEAs, we model the historical best-so-far objective values and predict the next best-so-far value. For MOEAs, where no unique best-so-far solution exists, we represent the quality of the evolving non-dominated solution set using an $R2$-based expected utility \cite{hansen1994evaluating}. 

The main contributions of this paper are threefold. First, we introduce a cost-aware stopping criterion for EAs that explicitly balances the predicted marginal improvement from continued search against the cost of additional function evaluations. Second, we propose a cost-aware performance metric to evaluate the stopping performance of both single- and multi-objective evolutionary optimization. Third, we evaluate the proposed stopping rule against representative existing termination criteria and a hindsight-best benchmark, allowing us to assess how closely an implementable online stopping rule approaches the best stopping decision that could have been selected with knowledge of the complete search trajectory. Experimental results show that our proposed criterion achieves a better trade-off between solution quality and function evaluation cost than existing stopping criteria, achieving performance close to the hindsight-best benchmark which has access to complete information.

\section{Benchmark stopping criteria \label{sec:preliminaries}}
We compare our proposed stopping criterion with a number of stopping criteria from the literature. 
For single-objective EAs, we chose \texttt{tolfun, tolfunhist, tolflatfiness} and \texttt{tolstagnation} which have been shown to work well for CMA-ES \cite{tanabe2026quantitative}. They all make stopping decisions based on the distribution of objective values in the offspring population, and how this distribution changes over time. Details can be found in the online appendix \footnote{The online appendix is available at \url{https://github.com/xinlione/cost-aware-stopping}.}.

For the MOEAs, our set of benchmarks comprises of MGBM \cite{marti2016stopping}, Entropy-based Stopping Criterion (ESC) \cite{saxena2015entropy}, and $\epsilon$-Progress-based Stopping Criterion ($\epsilon$SC) \cite{hadka2013borg} which all look at dominance relationships between the current population and previous populations. For more details, the reader is referred to the online appendix.

\section{Cost-aware Performance Measure} \label{sec:performance}
\subsection{Function evaluation cost}
Let $c_e$ denote the cost of one function evaluation, and $N$ be the number of function evaluations per iteration. At iteration $t$, the evaluation cost $C$ is $C = c_e N$. 
Note that we assume the function evaluation cost $c_e$  is expressed on the same scale as the solution performance, allowing the two terms to be directly combined. It can also be interpreted as objective-value units per function evaluation. 

\textcolor{black}{\subsection{Cost-aware Performance Measure} }
We define a unified cost-aware performance measure for both SOEAs and MOEAs. Let $\mathcal{S}^t$ denote the incumbent search output at generation $t$, and let $Q_t$ denote its scalar quality measure, where smaller values are preferred. 

\paragraph{SOEA.} We consider a single-objective minimization problem, i.e., $\min_x f(x)$. At generation $t$, let $\mathcal{P}_{\mathrm{SO}}^t$ denote the current population. The incumbent solution $\mathbf{x}_t$ is the best solution found up to generation $t$, and its best-so-far objective value is $B_t=f(\mathbf{x}_t)=\min_{s\leq t}\min_{x\in\mathcal{P}_{\mathrm{SO}}^s}f(x)$. Hence, $\{B_t\}$ forms a non-increasing best-so-far quality trajectory.

\paragraph{MOEA.} For a multi-objective minimization problem, let the objective vector of a solution $x$ be $\mathbf{f}(x)=(f_1(x),\ldots,f_m(x))$. At generation $t$, a MOEA yields a population $\mathcal{P}^t=\{x_1,\ldots,x_N\}$. Since there is no unique best solution in the multi-objective setting, we maintain an external non-dominated archive $\mathcal{A}^t$, defined as $\mathcal{A}^t=\operatorname{ND}(\mathcal{A}^{t-1}\cup\mathcal{P}^t)$, where $\operatorname{ND}(\cdot)$ retains only non-dominated solutions discovered up to generation $t$.

To evaluate the quality of $\mathcal{A}^t$, we use an expected-utility measure based on Tchebycheff scalarization with random preference weights, i.e., an $R2$-based metric \cite{hansen1994evaluating}. Let $\mathbf{w}=(w_1,\ldots,w_m)$ denote a preference vector satisfying $w_i\geq0$ and $\sum_{i=1}^m w_i=1$. For a given preference vector $\mathbf{w}$,
$d_{\mathbf{w}}(x)=\max_{i=1,\ldots,m}\{w_i(f_i(x)-z_i^{\mathrm{ideal}})\}$ denotes the distance of solution $x$ from the ideal point $\mathbf{z}^{\mathrm{ideal}}$, where a smaller value is preferred. The quality of the archive under preference $\mathbf{w}$ is therefore $D(\mathcal{A}^t,\mathbf{w})=\min_{x\in\mathcal{A}^t}d_{\mathbf{w}}(x)$. Let the preference vector $\mathbf{W}$ follow a uniform distribution over the unit simplex, i.e., $P_W=\operatorname{Dirichlet}(\mathbf{1}_m)$. We define $R_t=R2(\mathcal{A}^t)$ as $R_t=\mathbb{E}_{\mathbf{W}\sim P_W}\left[\min_{x\in\mathcal{A}^t}\max_{i=1,\ldots,m}\{W_i(f_i(x)-z_i^{\mathrm{ideal}})\}\right]$. The expectation under $P_W$ is approximated by Monte Carlo sampling. Specifically, we draw a fixed set of $U$ independent weight vectors $\mathbf{W}^{(u)}\sim\operatorname{Dirichlet}(\mathbf{1}_m)$, $u=1,\ldots,U$, at the beginning of each run and reuse the same weight vectors at every generation. We estimate $ R_t\approx\frac{1}{U}\sum_{u=1}^{U}\min_{x\in\mathcal{A}^t}\max_{i=1,\ldots,m}\{W_i^{(u)}(f_i(x)-z_i^{\mathrm{ideal}})\}$. A smaller $R_t$ indicates a better approximation set. 

The quality of the current population may deteriorate from one generation to the next, even for an elitist MOEA, since elitism does not necessarily guarantee monotonic improvement in the $R2$ indicator. Using the external non-dominated archive avoids this issue. Since $\mathcal{A}^{t-1}\subseteq\mathcal{A}^t$, its $R2$ value is non-increasing for the minimization formulation. Alternatively, one could simply retain the best-so-far population. In any case, $\{R_t\}$ provides a monotonic quality trajectory analogous to the best-so-far sequence $\{B_t\}$ in SOEAs.

To summarize, $(\mathcal{S}^t,Q_t)=(\mathbf{x}_t,B_t)$ for SOEAs, whereas $(\mathcal{S}^t,Q_t)=(\mathcal{A}^t,R_t)$ for MOEAs. Hence, the two settings differ only in how the incumbent search output and its quality are defined. Using the common quality measure $Q_t$, we define the cost-aware performance at termination generation $\tau$ as
\begin{align}
    \mathrm{CAP}_\tau = Q_\tau + \tau C. 
\end{align}

where $C$ denotes the cost of one generation. Accordingly, $\mathrm{CAP}^{\mathrm{SO}}_\tau=B_\tau+\tau C$ for SOEAs, while $\mathrm{CAP}^{\mathrm{MO}}_\tau=R_\tau+\tau C$ for MOEAs. A smaller $\mathrm{CAP}_\tau$ indicates a better trade-off between solution quality and evaluation cost.

\section{Cost-aware Stopping Criterion} \label{sec:criterion}
\subsection{Learning Curve Extrapolation}
Our goal is to predict the next best-so-far objective value by fitting a parametric model to the historical best-so-far trajectory and extrapolating it one iteration ahead. Existing parametric models of learning curves include a power law model with two and three parameters \cite{kadra2023scaling}, a logarithmic model, an exponential model \cite{swersky2014freeze}, the Morgan-Mercer-Flodin (MMF) model, and the Weibull model \cite{gu2001modelling}. No single model can describe all learning curves, therefore Domhan et al.~\cite{domhan2015speeding} propose a weighted linear combination of different models as the predictive model. We follow their method to build our learning curve model. 

We denote the historical best-so-far objective values up to $t$ by $\{B_s\}_{s=0}^{t}$ use our model to predict the next best-so-far objective value \(\hat{B}_{t+1}\). We consider a set of \(I=4\) parametric curve models \(\{g_1,\ldots,g_I\}\), where each model \(g_i\) is parameterized by a set of parameters \(\boldsymbol{\theta}_i\) at $t$, i.e., $g_i (t| \boldsymbol{\theta}_i)$.  They are chosen from the literature whose shapes align with our prior knowledge about the form of the learning curves, which are typically decreasing, saturating functions. Specifically, the candidate learning-curve models are exponential ($g_1(t) = a_1+A_1\exp(-\kappa_1 t)$), power law ($g_2(t) = a_2+A_2(t+b_2)^{-p_2}$), stretched exponential ($g_3(t) = a_3+A_3 \exp\!\left[-(\kappa_3 t)^{\beta_3}\right]$), and hyperbolic ($g_4(t) = a_4 + \frac{A_4}{(1+\kappa_4 t)^{p_4}}$). \(a_i\) represents the asymptotic objective value, \(A_i\) controls the initial distance from the asymptote, and the remaining parameters control the rate and shape of convergence. We impose \(A_i\geq0\), \(\kappa_i>0\), \(p_i>0\), \(\beta_3>0\), and \(b_2>0\), such that the models represent decreasing and saturating learning curves for a minimization problem.

We then combine these individual models into a single weighted linear combination model $ g_{\mathrm{com}}(t|\boldsymbol{\xi}) = \sum_{i=1}^I \omega_i g_i (t| \boldsymbol{\theta}_i)$, where the new combined parameter vector is $\boldsymbol{\xi} = (\omega_1, \ldots, \omega_I, \boldsymbol{\theta}_1, \ldots, \boldsymbol{\theta}_I)$, comprising a weight $\omega_i$ for each model and the individual model parameter $\boldsymbol{\theta}_i$. The historical best-so-far objective values \(B_s\) are approximated by the combined learning-curve model, $B_s \approx g_{\mathrm{comb}}(s\mid\boldsymbol{\xi})$ for $s=0,\ldots,t$. At generation $t$, the observed history $\{B_s\}_{s=0}^{t}$ is used to jointly estimate the model parameters and combination weights through the following deterministic constrained nonlinear least-squares problem:
\begin{align}
\min_{\boldsymbol{\xi}}  & \sum_{s=0}^{t} \left[ B_s- g_{\mathrm{comb}}(s|\boldsymbol{\xi}) \right]^2 \\
\text{s.t.} & \sum_{i=1}^{4} \omega_i=1, \label{eq:curve_weight_sum}\\
& \omega_i \geq 0, i=1,\ldots,4, \label{eq:curve_weight_nonnegative}\\
& 0 \leq g_{\mathrm{comb}}(t+1 | \boldsymbol{\xi}) \leq B_t. \label{eq:curve_prediction_constraint}
\end{align}

Constraints \eqref{eq:curve_weight_sum}-\eqref{eq:curve_weight_nonnegative} ensure that the fitted learning curve is a convex combination of the component models. Also, it enables us to interpret each model as a non-negative additive component of the learning curve. Constraint \eqref{eq:curve_prediction_constraint} incorporates the best-so-far property of a minimization problem: the predicted objective value after the next generation cannot be worse than the current best-so-far value. It ensures the non-increasing property of the best-so-far objective sequence in a minimization problem and non-negative predicted improvement. \textcolor{black}{We present an ablation study of the elitist prediction constraint in the online appendix.} The problem is solved using the Sequential Least Squares Programming (SLSQP) algorithm \cite{kraft1988software}, as it is a smooth nonlinear optimization problem with both bound and equality/inequality constraints. The model parameters and combination weights are estimated jointly, yielding the optimal parameter vector $\widehat{\boldsymbol{\xi}}_t = \arg\min_{\boldsymbol{\xi}} \sum_{s=0}^{t} \left[ B_s-g_{\mathrm{comb}}(s\mid\boldsymbol{\xi}) \right]^2$, subject to the constraints above. The fitted model is then used to obtain the one-step-ahead prediction of the next best-so-far objective value as 
\begin{align}
\widehat{B}_{t+1} = g_{\mathrm{com}} (t+1\mid\hat{\boldsymbol{\xi}}_t). \label{eq:prediction}
\end{align}

The model is re-estimated after each generation as new observations become available, allowing both the component parameters and their combination weights to adapt to the observed search trajectory.

\subsubsection{An illustrative prediction example.} To illustrate the learning-curve fitting and prediction procedure, we consider a representative CMA-ES run on one \texttt{bbob} problem $f_{15}$ \cite{hansen2009real} under the same SOEA experimental setting described in Section~\ref{sec:setup}. At a selected generation $t$, only the historical best-so-far objective values $\{B_s\}_{s=0}^{t}$ are used to estimate the parameters and combination weights of the learning-curve model. The fitted model is then extrapolated by one generation to obtain the prediction $\widehat{B}_{t+1}$. For illustration, the subsequently observed value $B_{t+1}$ is also reported to demonstrate the one-step prediction quality. 

\begin{figure}[tb]
\centering
\includegraphics[width=0.7\textwidth]{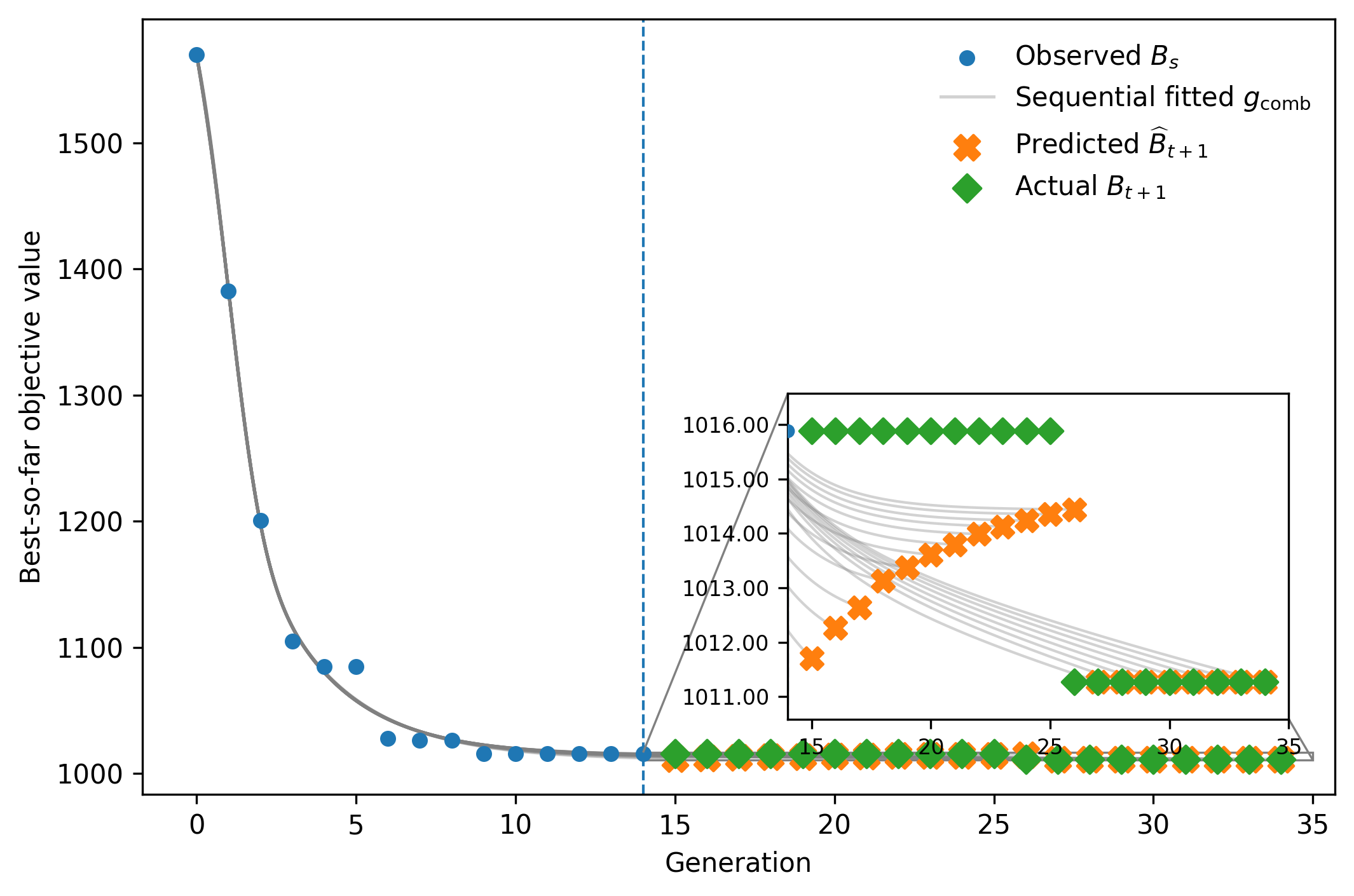}
\caption{An illustrative prediction example of a representative CMA-ES run for one instance of $f_{15}$ in \texttt{bbob} functions, dimensions $n=3$. The prediction starts at generation $15$ and iterates for the following 20 generations. The weights for the last fitted curve at generation 35 are $\omega_1 = 0.62, \omega_2 = 0.00, \omega_3 = 0.35, \omega_4 = 0.03$.}  
\label{fig:regression_example}
\end{figure}

From Figure~\ref{fig:regression_example}, the best-so-far objective value decreases rapidly in the early generations, followed by a long plateau and smaller subsequent improvements. Starting from generation 15, the model produces predictions $\widehat{B}_{t+1}$ that are generally close to the realized values $B_{t+1}$. When a sharp improvement occurs around generation 26, the prediction quickly adapts to the new objective level and subsequently tracks the best-so-far trajectory closely. Overall, the combined learning-curve model captures the convergence pattern well and provides reliable one-step-ahead predictions.

\subsection{Cost-aware Stopping Criterion} 

Using the unified quality measure $Q_t$ introduced above, We propose a unified cost-aware stopping criterion for both single- and multi-objective evolutionary algorithms since the two settings differ only in how the incumbent search output and its quality are defined; the prediction and stopping mechanism is identical. 

At generation $t$, we fit the learning-curve model $g_{\mathrm{comb}}(s\mid\boldsymbol{\xi})$ to the historical quality sequence $\{Q_s\}_{s=0}^{t}$ and predict the next-generation quality $\widehat{Q}_{t+1}$. The predicted one-generation improvement is $ \widehat{\Delta Q}_t = Q_t-\widehat{Q}_{t+1}$. Let $C$ denote the cost of one additional generation.  The search is continued only when the predicted improvement exceeds this cost, i.e., $\widehat{\Delta Q}_t>C$; otherwise, the stopping condition is triggered.

Directly applying the stopping condition, however, may lead to premature stopping. In the early stage of the search, only a small number of observations are available, and the fitted learning curve may not yet provide a reliable prediction of future progress. We therefore introduce a warm-up period of \(T_{\min}\) generations, where $T_{\min}$ is the minimum number of generations before stopping is allowed. This provides the prediction model with sufficient search history before it is used for termination decisions, and avoids premature stopping caused by unstable early predictions. 
Moreover, evolutionary search is stochastic and may exhibit temporary stagnation followed by further improvement. A single generation with a small predicted improvement is therefore not sufficient evidence of convergence. To make the stopping decision less sensitive to short-term fluctuations and prediction error, we require the stopping condition to hold for $M$ consecutive generations. Specifically, the search terminates when
\begin{align}
    \widehat{\Delta Q}_{t-j} \leq C_{t-j}, j=0,\ldots,M-1. \label{eq:soea_stop}
\end{align}

Therefore, $\tau = \inf \left\{ t\ge T_{\min}+M-1: \widehat{\Delta Q}_{j} \leq C_{j}, \forall j=t-M+1,\ldots,t \right\}$. The procedure is summarized in Algorithm~\ref{alg:cost_aware_stopping}.

\begin{algorithm}[!htbp]
\caption{EA with the proposed cost-aware stopping criterion}
\label{alg:cost_aware_stopping}
\begin{algorithmic}[1]
\State Initialize and evaluate $\mathcal{P}^{0}$; construct $\mathcal{S}^{0}$ and compute $Q_0$
\State $t\gets0$, $k\gets0$
\While{$t<t^{\max}$}
    \State Generate and evaluate $\mathcal{P}^{t+1}$ using the EA operators
    \State Update $\mathcal{S}^{t+1}$ and compute $Q_{t+1}$
    \State $t\gets t+1$
    \If{$t\geq T_{\min}$}
        \State Fit $g_{\mathrm{comb}}$ to $\{Q_s\}_{s=0}^{t}$ and predict $\widehat{Q}_{t+1}$
        \State $\widehat{\Delta Q}_t\gets Q_t-\widehat{Q}_{t+1}$;
        $C \gets c_e N$
        \State $k\gets k+1$ if $\widehat{\Delta Q}_t\leq C$, otherwise $k\gets0$
        \If{$k\geq M$}
            \State \textbf{break}
        \EndIf
    \EndIf
\EndWhile
\State \Return $\mathcal{S}^{t}$
\end{algorithmic}
\end{algorithm}


\section{Experiment Setup} \label{sec:setup}
\subsubsection{Setting for Single-Objective Problem.} 
We evaluate the proposed and existing stopping criteria on the noiseless \texttt{bbob} benchmark suite \cite{hansen2009real}. 
We consider dimensions $n\in\{2,3,5\}$ and perform 15 independent runs for each function following the  Comparing Continuous Optimizers (COCO) framework \cite{hansen2021coco}. The optimum of each \texttt{bbob} instance lies within $[-5,5]^n$. 
We use CMA-ES as the search algorithm, as it is a state-of-the-art evolutionary algorithm for continuous black-box optimization \cite{hansen2016cma}. At each generation, CMA-ES produces a population of candidate solutions and their objective values, from which we record the best-so-far objective value $B_t$.

All experiments use CMA-ES \cite{hansen2001completely} implemented in \texttt{pycma} \cite{hansen2024cma}. The offspring population size is $\lambda = 4 + \lfloor 3\ln(n) \rfloor$, the initial solution is sampled uniformly from $[-4,4]^n$ \cite{hansen2009benchmarking}, and the initial step size is $\sigma_0=2$. Other CMA-ES and stopping-criterion parameters use their default settings. The maximum evaluation budget is $\mathrm {FE}_{\max} = 10^4 \times n$. Each run continues until this budget is exhausted,
regardless of whether a stopping criterion has already triggered or the optimum has been reached. The search also continued after reaching the optimum, allowing us to assess whether the stopping criteria could correctly detect convergence. For the stopping criteria, we set $\epsilon_{tfun}=10^{-11}$, $\epsilon_{thist}=10^{-12}$. $T_{\mathrm{tstag}}=\lfloor 100 + 100 \times n \times 1.5/\mu \rfloor$ by default. 

To enable comparison across \texttt{bbob} functions with different objective scales, we normalize the best-so-far objective value using a reference initial optimality gap. For each function $f$ and dimension $n$, we generate 1,000 independent initial CMA-ES populations using the same initialization procedure as in the main experiments. For each initialization run, we compute the initial optimality gap $G_0^{(r)}=B_0^{(r)}-f_{\mathrm{opt}}^{(r)}$, where $B_0^{(r)}$ is the best objective value in the initial population and \(f_{\mathrm{opt}}^{(r)}\) is the known optimum of the corresponding \texttt{bbob} instance. With the average reference initial gap $\bar G_0(f,n) =\frac{1}{1000}\sum_{r=1}^{1000}G_0^{(r)}$, we define the normalized value at generation $t$ as $L_t=\frac{B_t - f_{\mathrm{opt}}}{\bar G_0(f,n)}$. Thus, $L_0 \approx 1$ and $L_t$ approaches zero as it reaches the optimum. We then define the unit function-evaluation cost as $c_e= \rho /FE_{\max}$, where $\rho$ controls the relative importance of evaluation cost over solution quality. The proposed rule stops when $L_t-\widehat{L}_{t+1}\leq \rho\lambda/FE_{\max}$. The cost-aware performance at stopping generation $\tau$ is $ \mathrm{CAP}^{\mathrm{SO}}_\tau = L_\tau+\rho\frac{\sum_{i=1}^{\tau}N_i}{FE_{\max}}$, where a smaller value indicates a better trade-off between solution quality and evaluation cost. Unless specified otherwise, we set $\rho=10$, $T_{\min}=50$ generations, and the patience parameter $M=5$ generations.

\subsubsection{Setting for Multi-Objective Problem.} 
We evaluate the proposed cost-aware stopping rule on DTLZ1-DTLZ7 \cite{deb2005scalable}, with \(m\in\{2,3,5\}\) objectives using MOEA/D \cite{zhang2007moea}. The population size is fixed at 50, and the remaining MOEA/D settings follow the default \texttt{pymoo} \cite{blank2020pymoo} implementation, using Tchebycheff decomposition for \(m=2\) and PBI for \(m=3,5\). Each configuration is repeated 31 times with a maximum budget of \(\mathrm {FE}_{\max} =10^4\) function evaluations. An external non-dominated archive is maintained, and objective vectors are normalized using the true ideal and nadir points before computing the \(R2\) indicator with 1,000 fixed preference vectors. For the benchmark criteria, MGBM uses \(R=0.05\) with a stopping threshold of \(0.12\); the entropy-based criterion uses a 20-generation window, 10 bins, and a rounding precision of \(0.001\); and the \(\epsilon\)-progress criterion uses \(\epsilon=0.2\) with a stagnation limit of 10 generations \cite{kitamura2026benchmarking}. Performance is evaluated by the cost-aware performance metric, \(CAP^{MO}_t\). We set $T_{\min} = 20, \rho = 10$. Unless otherwise specified, the parameter settings are the same as those used for the SOEA experiments.

\section{Results} \label{sec:results}
For performance comparison, we include a hindsight-best benchmark as an ideal reference on the same complete trajectories in both experiments. It retrospectively selects the iteration that achieves the minimum cost-aware performance over the complete search trajectory. Since it uses future information unavailable to an online stopping rule, it cannot be implemented in practice. 

\subsection{Stopping in SOEAs}
We first present the average stopping performance of the different criteria, together with the hindsight-best benchmark in Figure~\ref{fig:soea_performance}. Results for $n=3$ and $n=5$ are provided in the appendix. 
 
\begin{figure}[!htb]
\centering
\includegraphics[width=0.8\textwidth]{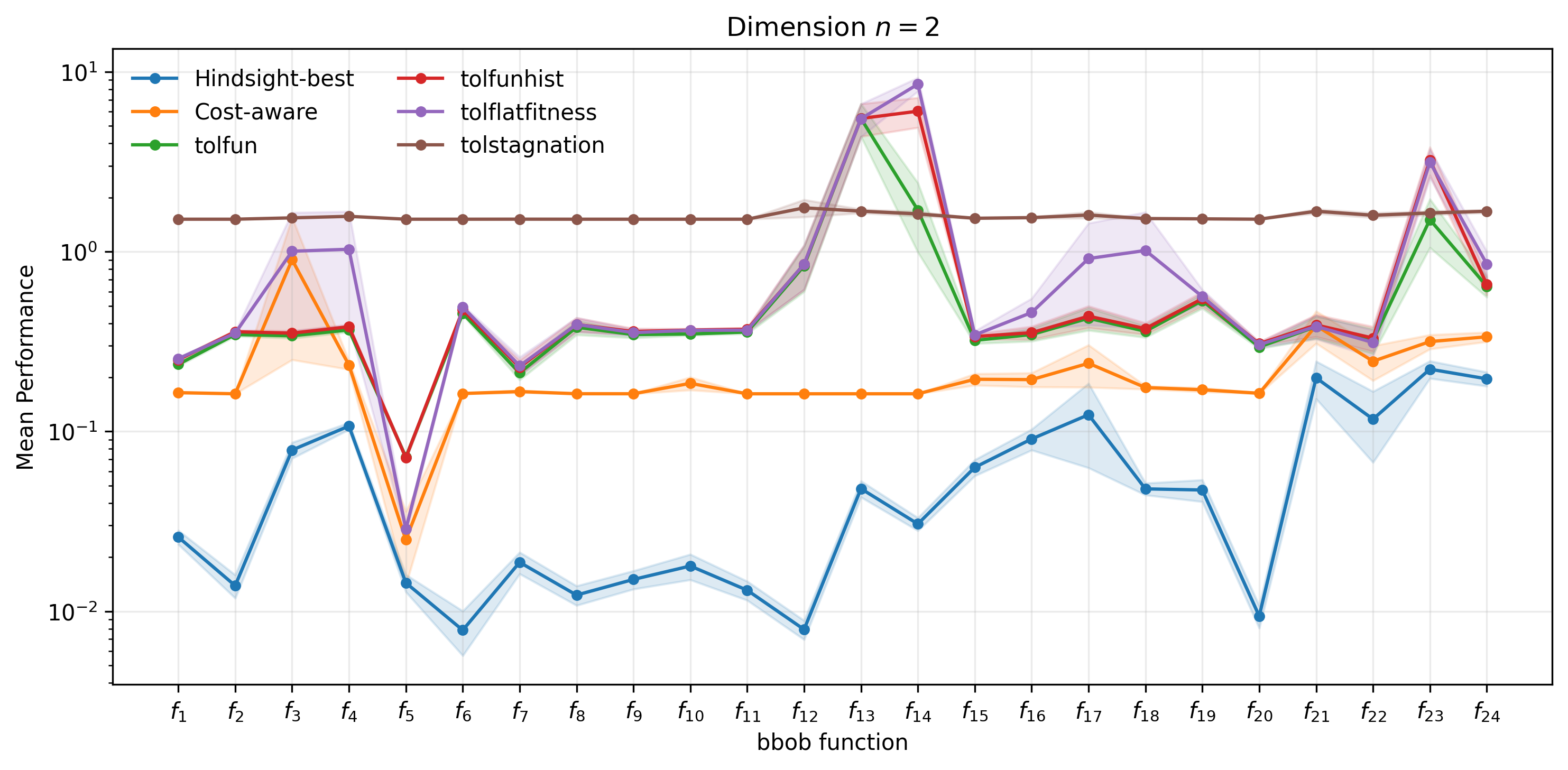}
\caption{Mean cost-aware performance of the stopping methods across the 24 \texttt{bbob} functions for dimensions $n=2$. Solid lines shows the mean performance over 15 problem instances, and the shaded region represents $\pm$ one standard error. Lower values indicate better stopping performance. } 
\label{fig:soea_performance}
\end{figure}

\begin{table}[!htb]
\caption{Mean cost-aware performance of the stopping criteria across the \texttt{bbob} functions. Results are reported as mean $\pm$ standard error over the problem instances. Lower values indicate better performance.}
\label{tab:soea_performance_summary}
\centering
\begin{tabular}{l C{3cm} C{3cm} C{3cm} }
\toprule
\textbf{Criterion} & \textbf{$n=2$} & \textbf{$n=3$} & \textbf{$n=5$} \\
\midrule
\texttt{tolfun}         & $0.69 \pm 0.08$ & $1.27 \pm 0.14$ & $1.60 \pm 0.16$ \\
\texttt{tolfunhist}     & $0.96 \pm 0.11$ & $1.52 \pm 0.16$ & $2.02 \pm 0.18$ \\
\texttt{tolflatfitness} & $1.17 \pm 0.13$ & $1.96 \pm 0.18$ & $2.57 \pm 0.21$ \\
\texttt{tolstagnation}  & $1.57 \pm 0.01$ & $1.37 \pm 0.05$ & $1.08 \pm 0.06$ \\
Cost-aware              & $\textbf{0.22} \pm 0.03$ & $\textbf{0.26} \pm 0.04$ & $\textbf{0.19} \pm 0.03$ \\
Hindsight-best*         & $0.06 \pm 0.01$ & $0.07 \pm 0.00$ & $0.08 \pm 0.00$ \\
\bottomrule
\end{tabular}
\end{table}

\begin{table}[!htb]
\centering
\caption{Mean function evaluations across \texttt{bbob} problem instances for each stopping criterion. Results are reported as mean $\pm$ standard error over the problem instances. Lower values indicate early stopping.}
\label{tab:soea_stopping_summary}
\begin{tabular}{l C{3cm} C{3cm} C{3cm}}
\toprule
\textbf{Criterion} & \textbf{$n=2$} & \textbf{$n=3$} & \textbf{$n=5$}  \\
\midrule
\texttt{tolfun}          & $1331.08 \pm 161.62$ & $3747.28 \pm 421.70$ & $7881.13 \pm 809.92$  \\
\texttt{tolfunhist}      & $1859.60 \pm 216.48$ & $4494.47 \pm 469.53$ & $9984.18 \pm 921.93$  \\
\texttt{tolflatfitness}  & $2282.35 \pm 254.78$ & $5824.16 \pm 538.03$ & $12728.73 \pm 1031.67$ \\
\texttt{tolstagnation}   & $3081.57 \pm 17.77$  & $4042.13 \pm 162.74$ & $5243.51 \pm 299.52$  \\
Cost-aware               & $\textbf{389.37} \pm 55.09$    & $\textbf{676.74} \pm 125.04$  & $\textbf{738.24} \pm 145.16$   \\
Hindsight-best*          & $50.40 \pm 2.43$     & $89.08 \pm 4.47$     & $185.84 \pm 9.88$     \\
\bottomrule
\end{tabular}
\end{table}

From Figure \ref{fig:soea_performance}, we first observe that for most \texttt{bbob} functions, the stopping criteria show relatively similar performance, but with our cost-aware stopping rule always performing best. Larger differences appear for $f_{13}$, $f_{14}$, and $f_{23}$. On these functions, the search can continue for many generations with only small or slow improvements, so criteria such as \texttt{tolfun}, \texttt{tolfunhist}, and \texttt{tolflatfitness} are not triggered at all for some instances and the run continues until $FE_{\max}$. Since the proposed performance measure also accounts for function-evaluation cost, these long runs lead to noticeably higher performance values than criteria that stop earlier.  With the overall performance results in Table~\ref{tab:soea_performance_summary}, we can see that its overall mean performance at $n=2$ is $0.22$, compared with $0.69$ to $1.57$ for the other existing criteria. The hindsight-best benchmark achieves the best overall mean of $0.06$ as expected because it selects the stopping point using the complete search trajectory. The relatively small gap between the proposed rule and hindsight-best indicates that the proposed rule can identify a near-desirable stopping point without access to future information. Finally, the existing criteria show substantial variation across \texttt{bbob} functions, whereas the proposed rule is comparatively more stable. Although its performance value changes from $0.22$ at $n=2$ to $0.19$ at $n=5$, the proposed rule remains the best-performing implementable criterion across all three dimensions. 

Table~\ref{tab:soea_stopping_summary} reports the mean number of function evaluations at termination. The proposed cost-aware criterion stops substantially earlier than the four existing benchmark criteria across all dimensions, with the difference becoming more pronounced as the dimension increases. Moreover, the stopping points of the cost-aware rule are close to the hindsight-best benchmark, which suggests that our proposed stopping rule can identify a stopping point close to the ideal retrospective benchmark while using only information available during the search.

\subsection{Stopping in MOEAs}
We first present the average stopping performance of the different criteria, together with the hindsight-best benchmark, for $m=3$ in Figure~\ref{fig:moea_performance}. Results for $m=2,5$ are provided in the appendix. 

\begin{figure}[tb] 
\centering
\includegraphics[width=0.8\textwidth]{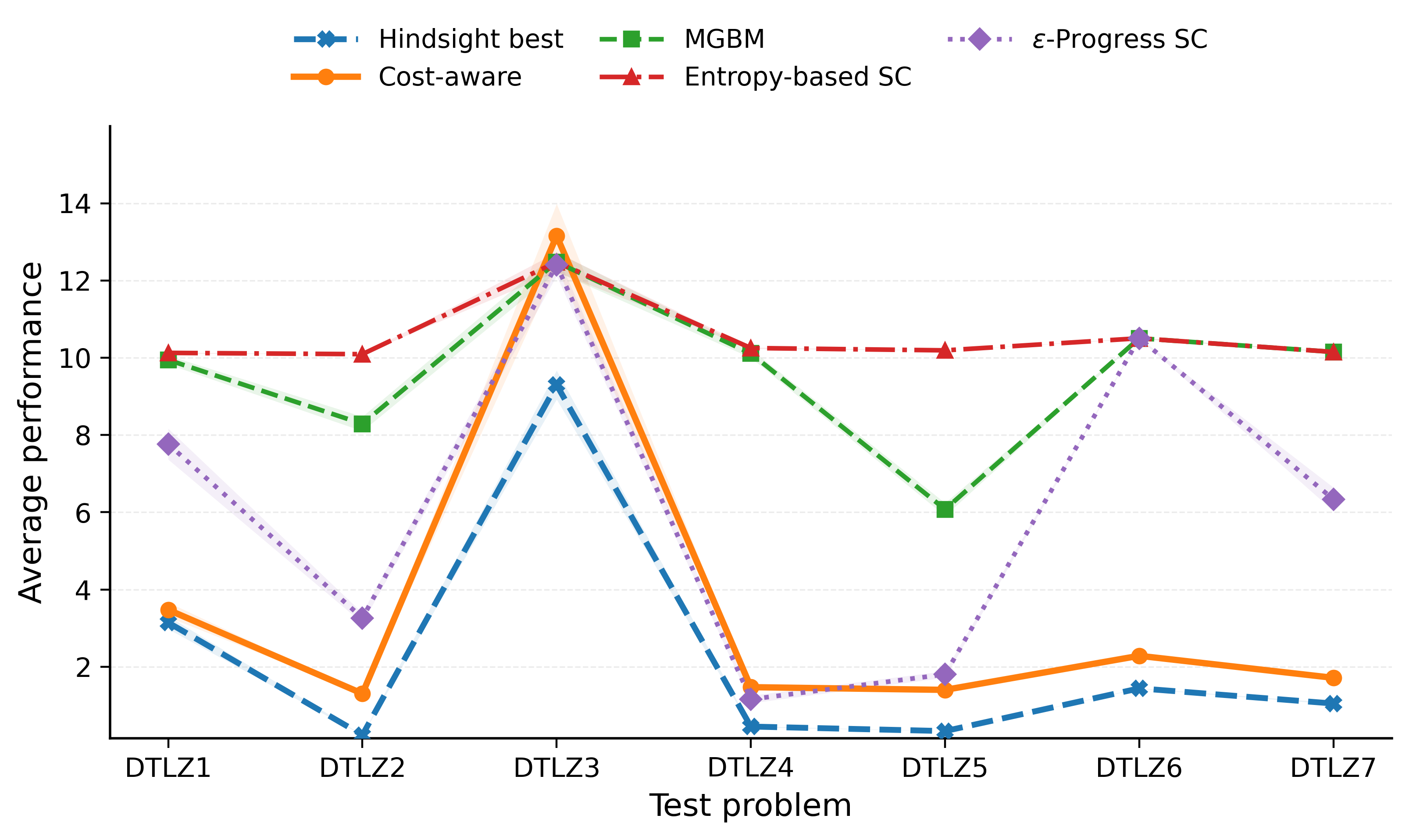}
\caption{Mean normalized cost-aware performance of the stopping criteria across DTLZ1–DTLZ7 problems for 3 objectives. The line shows the mean performance over 31 replications, and the shaded region represents $\pm$ one standard error. Lower values indicate better stopping performance.} 
\label{fig:moea_performance}
\end{figure}

\begin{table}[htb]
\centering
\caption{Mean cost-aware performance of the stopping criteria across the DTLZ1-DTLZ7 problems and all 31 runs. Results are reported as mean $\pm$ standard error over the problem instances. Lower values indicate better performance.} 
\label{tab:moea_performance_summary}
\begin{tabular}{l C{3cm} C{3cm} C{3cm}}
\toprule
\textbf{Criterion} & \textbf{$m=2$} & \textbf{$m=3$} & \textbf{$m=5$} \\
\midrule
MGBM                   & $10.44 \pm 0.04$ & $9.65 \pm 0.14$ & $6.56 \pm 0.24$ \\
Entropy-based SC       & $10.44 \pm 0.04$ & $10.54 \pm 0.07$ & $10.30 \pm 0.03$ \\
$\epsilon$-Progress SC & $4.63 \pm 0.24$  & $6.18 \pm 0.29$ & $6.43 \pm 0.26$ \\
Cost-aware             & $\textbf{4.03} \pm 0.32$  & $\textbf{3.54} \pm 0.30$ & $\textbf{2.64} \pm 0.19$ \\
Hindsight-best         & $2.52 \pm 0.19$ & $2.28 \pm 0.21$ & $1.64 \pm 0.16$ \\
\bottomrule
\end{tabular}
\end{table}

\begin{table}[htb]
\centering
\caption{Mean function evaluations over all DTLZ1–DTLZ7 problems and all 31 runs for each stopping criterion. Results are reported as mean $\pm$ standard error over the problem instances. Lower values indicate early stopping.} 
\label{tab:moea_stopping_summary}
\begin{tabular}{l C{3cm} C{3cm} C{3cm}}
\toprule
\textbf{Criterion} & \textbf{$m=2$} & \textbf{$m=3$} & \textbf{$m=5$} \\
\midrule
MGBM                   & $10000.00 \pm 0.00$ & $9107.83 \pm 106.83$ & $6076.50 \pm 245.68$ \\
Entropy-based SC       & $10000.00 \pm 0.00$ & $10000.00 \pm 0.00$ & $10000.00 \pm 0.00$ \\
$\epsilon$-Progress SC & $4134.79 \pm 220.41$ & $5556.45 \pm 249.24$ & $6091.71 \pm 247.77$ \\
Cost-aware             & $\textbf{1329.95} \pm 26.38$ & $\textbf{1289.63} \pm 29.83$ & $\textbf{1275.58} \pm 17.81$ \\
Hindsight-best         & $1197.93 \pm 129.06$ & $874.88 \pm 111.48$ & $695.85 \pm 94.31$ \\
\bottomrule
\end{tabular}
\end{table}

From Figure~\ref{fig:moea_performance}, we can see that MGBM and entropy-based SC show poor performance across DTLZ1 to DTLZ7. The performance of $\epsilon$-Progress varies substantially across problems. Our proposed adaptive cost-aware criterion remains the closest to the hindsight-best benchmark across most problems and stays relatively stable performance across DTLZ1 to DTLZ7 (except DTLZ3), indicating a balance between the predicted improvement against evaluation cost. \textcolor{black}{DTLZ3 is particularly challenging for all stopping criteria, including the hindsight-best benchmark due to its highly multi-modal landscape.}


Table~\ref{tab:moea_performance_summary} further confirms the strong overall performance of the proposed cost-aware stopping rule. It achieves the best performance among the implementable criteria, which also remains relatively close to the hindsight-best benchmark, whereas MGBM and the entropy-based criterion have substantially larger values. From Table~\ref{tab:moea_stopping_summary}, we can see that MGBM is not triggered for $m=2$, while the entropy-based SC is not triggered for any objective setting within the budget and therefore stops at the maximum number of function evaluations. This may indicate that these criteria continue to detect ongoing search progress and therefore fail to terminate within the evaluation budget, leading to poorer cost-aware performance, as shown in Table~\ref{tab:moea_performance_summary}. Furthermore, we can see that MGBM typically runs close to the maximum budget, while $\epsilon$-Progress requires more evaluations as the number of objectives increases. In contrast, the proposed cost-aware rule stops consistently at around 1,300 FEs across all objective settings, which are closer to the hindsight-best stopping points than those of the other implementable criteria.


\subsection{Sensitivity Analysis of Cost}
To examine the sensitivity of the stopping performance to the evaluation-cost parameter, we consider four values of $\rho = \{0.1, 1, 10, 20\}$, representing increasingly expensive cost on function evaluations.  For the SOEA, we conduct the analysis on the \texttt{bbob} benchmark with dimension $n=2$. For the MOEA, we conduct the same analysis on the DTLZ1 to DTLZ7 benchmark problems with $m=3$ objectives. All other experimental settings remain the same as above. 


\begin{figure*}[tb]
\centering
\begin{subfigure}[t]{0.49\textwidth}
    \centering
    \includegraphics[width=\linewidth]{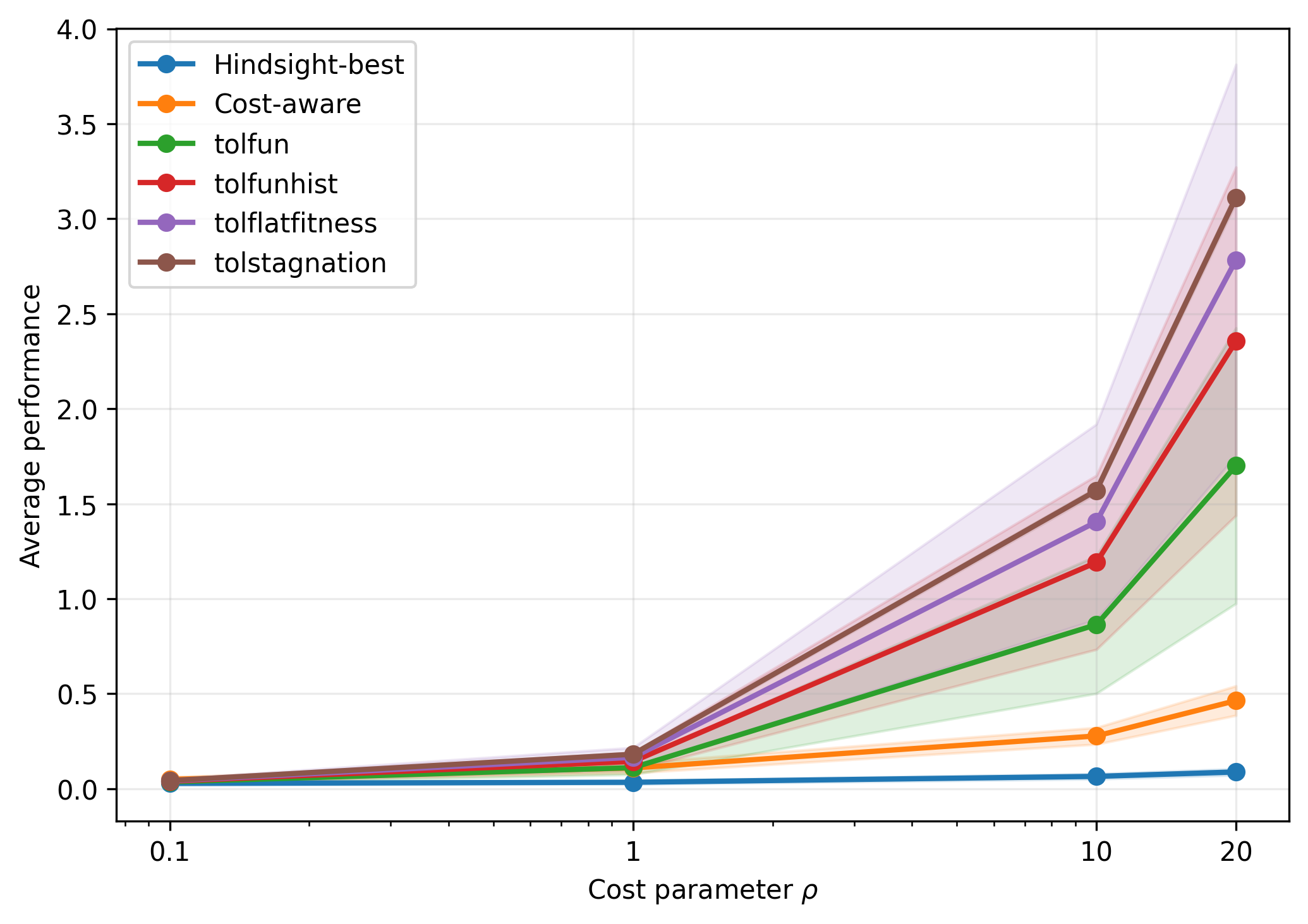}
    \caption{SOEA: cost-aware performance.}
    \label{fig:soea_cost_performance}
\end{subfigure}
\hfill
\begin{subfigure}[t]{0.49\textwidth}
    \centering
    \includegraphics[width=\linewidth]{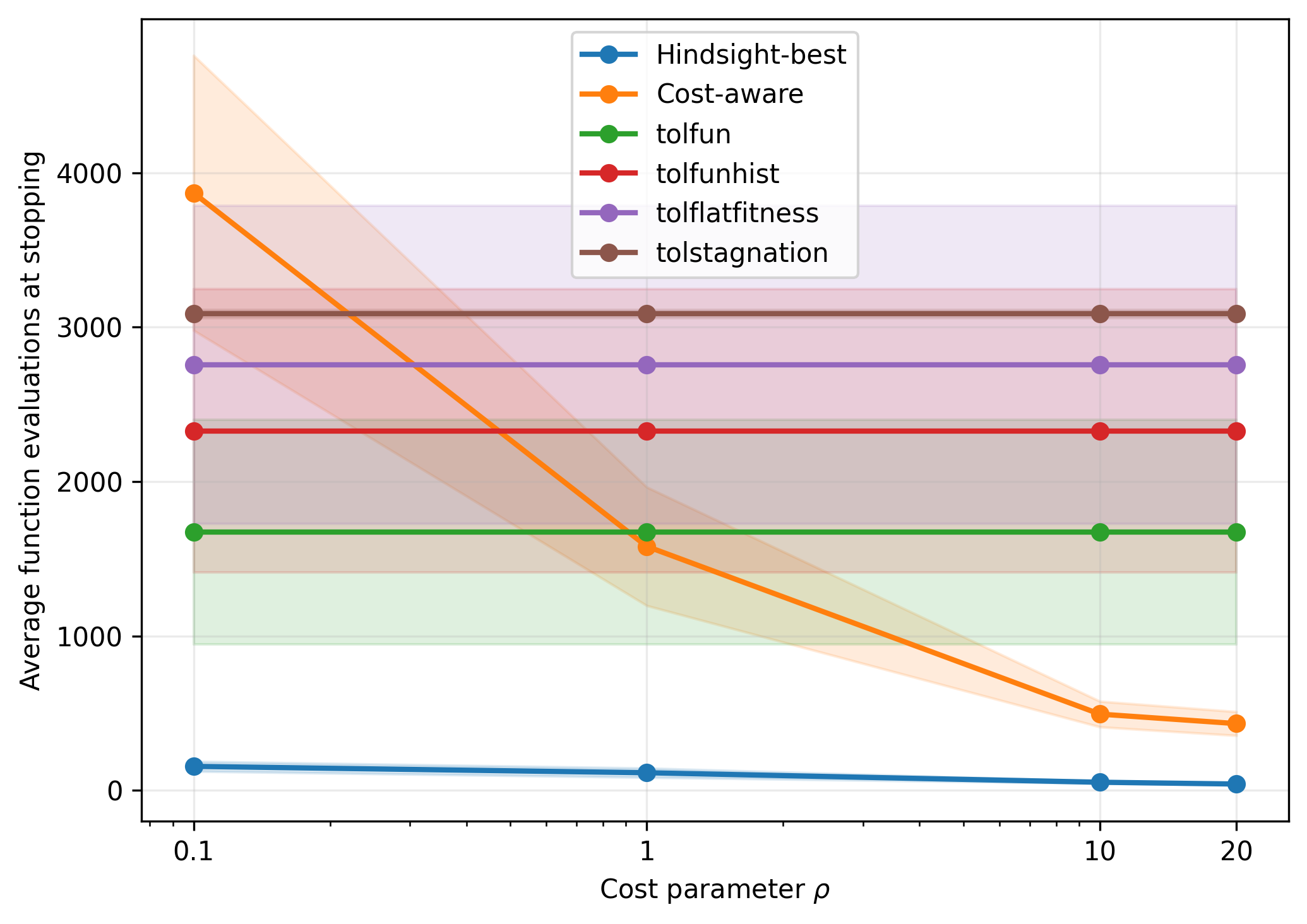}
    \caption{SOEA: stopping FEs.}
    \label{fig:soea_cost_stop}
\end{subfigure}

\vspace{0.5em}

\begin{subfigure}[t]{0.49\textwidth}
    \centering
    \includegraphics[width=\linewidth]{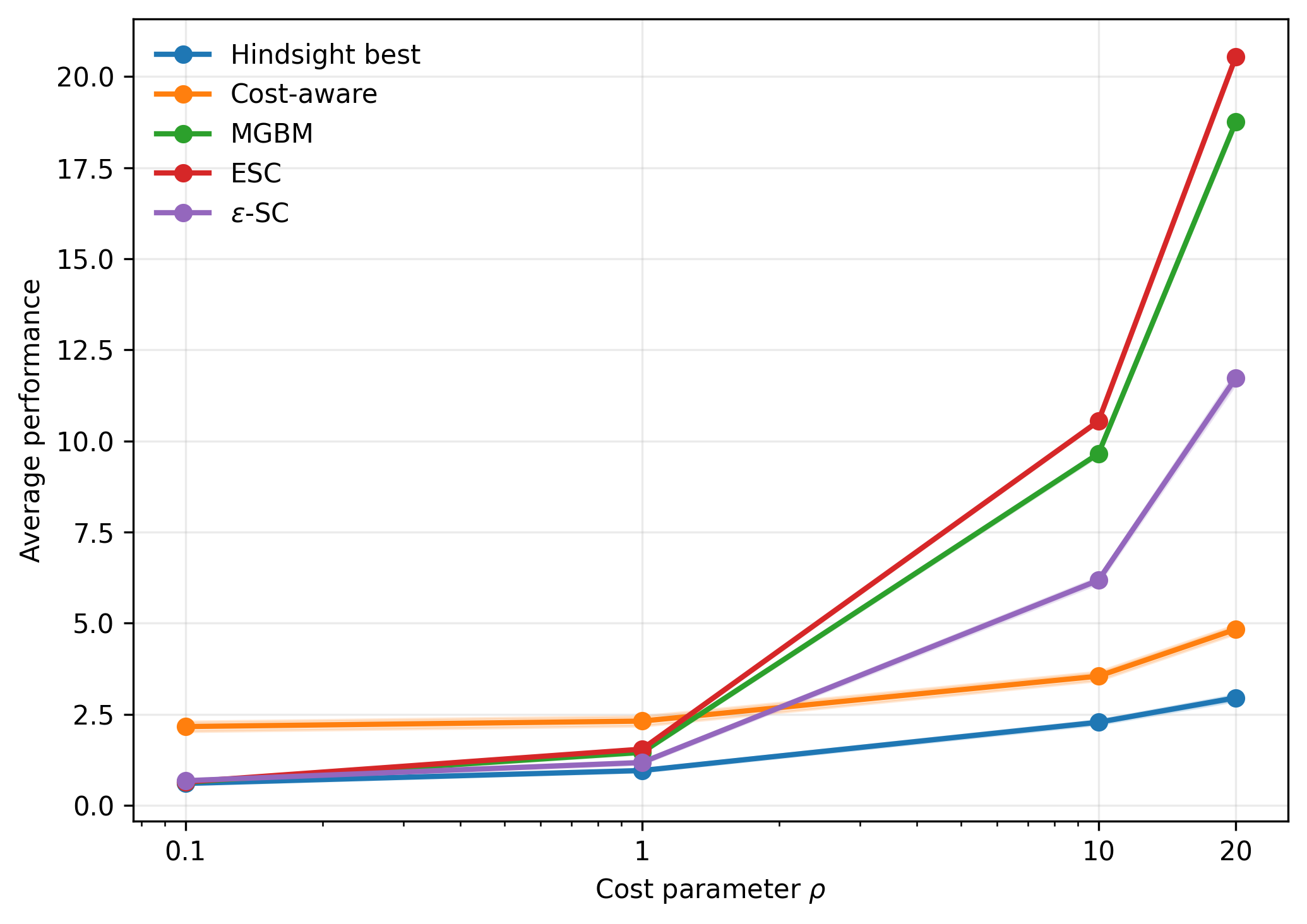}
    \caption{MOEA: cost-aware performance.}
    \label{fig:moea_cost_performance}
\end{subfigure}
\hfill
\begin{subfigure}[t]{0.49\textwidth}
    \centering
    \includegraphics[width=\linewidth]{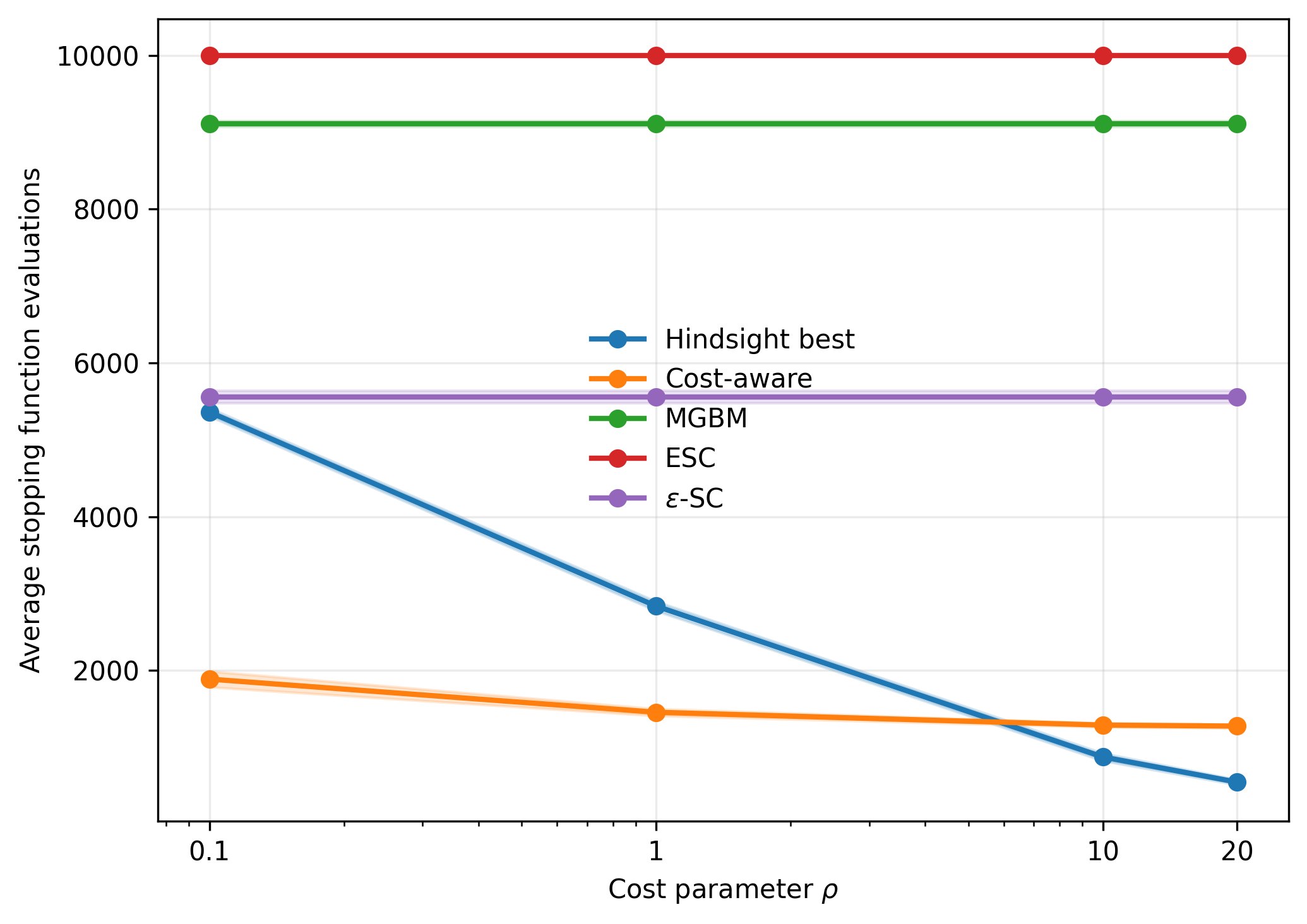}
    \caption{MOEA: stopping FEs.}
    \label{fig:moea_cost_stop}
\end{subfigure}

\caption{Sensitivity to the function-evaluation cost parameter for SOEA ($n=2$) and MOEA ($m=3$). Lines indicate the mean and shaded regions represent $\pm$ one standard error.}
\label{fig:sensitivity_cost}
\end{figure*}

Several observations can be made from Figure~\ref{fig:soea_cost_performance}. The average performance value increases with the cost parameter, as the evaluation-cost term becomes more heavily weighted in the performance metric. The cost-aware criterion remains among the best-performing methods across all cost settings. When $\rho$ is small, its performance is similar to that of the other criteria, whereas the performance gap becomes more pronounced as $\rho$ increases, as only our proposed stopping criterion can adapt to the higher cost and stop earlier (Figure~\ref{fig:soea_cost_stop}). 


From the results of the MOEA, the cost parameter $\rho$ has a clear effect on both performance and stopping behavior from Figure~\ref{fig:moea_cost_performance} and Figure~\ref{fig:moea_cost_stop}. When $\rho$ is small, evaluation cost is negligible and performance is mainly determined by solution quality. The cost-aware criterion stops relatively early and may therefore sacrifice useful later R2 improvements, resulting in worse performance than the other criteria. As $\rho$ increases, evaluation cost becomes more important and the differences become more pronounced. Since MGBM, ESC, and $\epsilon$-SC do not adapt their stopping decisions to $\rho$, their long runs incur increasingly large costs, whereas the cost-aware criterion stops earlier and consequently achieves better performance among the implementable criteria when $\rho$ increases. The hindsight-best benchmark shows a stronger response to $\rho$: it stops later when evaluations are cheap but much earlier when they are expensive. 

We observe different hindsight-best behavior in SOEA and MOEA: the average stopping FEs is relatively stable in SOEA but decreases markedly in MOEA as $\rho$ increases. This is mainly due to different convergence profiles. In SOEA, most normalized objective improvement occurs early, so the hindsight-optimal stopping point remains near the beginning of the search. In MOEA, meaningful R2 improvement continues over a larger fraction of the evaluation budget; hence, when evaluations are cheap, continuing is worthwhile, while larger $\rho$ shifts the optimum earlier. The difference is also affected by scaling. The SOEA loss is normalized toward zero, whereas the MOEA R2 indicator is not restricted to the same numerical range. In addition, $\mathrm{FE}_{\max}=10^4 \times n$ in SOEA but $10^4$ in MOEA. Therefore, the same $\rho$ represents different relative trade-offs between solution quality and evaluation cost in the two settings.


\section{Conclusion} \label{sec:conclusion}
This paper proposed a cost-aware stopping framework for both single- and multi-objective EAs. The proposed rule predicts the benefit of further search and terminates when the predicted improvement no longer justifies the cost of additional function evaluations. We also introduced a cost-aware performance metric that jointly considers solution quality and evaluation cost. Experiments on SOEA and MOEA settings show that the proposed rule achieves a favorable trade-off between search quality and computational cost, and performs close to the oracle stopping benchmark with hindsight information. 
Our results are preliminary and may be sensitive to parameter settings. Future work could investigate uncertainty-aware prediction models, adaptive estimation of evaluation cost, and validation across a broader range of EAs, particularly in  real-world optimization problems.

\bibliographystyle{splncs04}
\bibliography{references}

\appendix
\setcounter{figure}{0}
\renewcommand{\thefigure}{\thesection.\arabic{figure}}
\setcounter{table}{0}
\renewcommand{\thetable}{\thesection.\arabic{table}}

\noindent\textbf{\textit{This online appendix is provided as supplementary material and will not be included for publication.}}

\section{Benchmark Stopping criteria}
\subsection{Benchmark stopping criteria for SOEAs}
\paragraph{\texttt{tolfun}} makes a stopping decision based on the fitness values. At iteration $t$, the minimum and maximum objective values among the $\lambda$ offspring are defined as: $F_{\min}^t = \min_{i \in \{1, \ldots, \lambda\}} f(x_i^t)$ and $F_{\max}^t = \max_{i \in \{1, \ldots, \lambda\}} f(x_i^t)$, respectively. A set of historical minimum objective values over the most recent $l$ iterations is defined as $\mathcal H^t = \{F_{\min}^t, \ldots, F_{\min}^{t-l}\}$, where $l=10+ \lfloor 30n/\lambda \rfloor$, and $n$ is the dimension of the search space. The search stops when both $F_{\max}^t - F_{\min}^t < \epsilon_{tfun}$ and $\max \mathcal H^t - \min \mathcal H^t < \epsilon_{tfun}$ are met. 

\paragraph{\texttt{tolfunhist}} stops the search when $\max \mathcal H^t - \min \mathcal H^t < \epsilon_{thist}$. Different from \texttt{tolfun}, it ignores the condition on the difference between the minimum and maximum objective values.

\paragraph{\texttt{tolflatfitness}.} Define $F^t_{\lfloor\alpha \lambda \rfloor}$ as the $\alpha$-th percentile of the objective values among the $\lambda$ offspring at iteration $t$. 
The tolflatfitness criterion stops the search when $F_{\min}^t = F^t_{\lfloor 0.75 \lambda \rfloor}$ for $T_{\text{tflat}}$ consecutive iterations. By default, the $T_{\text{tflat}} = 1$ in \texttt{pycma}. Therefore, the search is stopped immediately when $F_{\min}^t = F^t_{\lfloor 0.75 \lambda \rfloor}$. 

\paragraph{\texttt{tolstagnation}} detects whether CMA-ES has stopped making meaningful progress. Let $\mathcal{L}$ and $\mathcal{M}$ denote the sets of minimum and median objective values among the $\lambda$ offspring, respectively, recorded every five iterations. For example, when $t=20$, \( \mathcal{L}^{(t)} = \left\{ F_{\min}^{(5)}, F_{\min}^{(10)}, F_{\min}^{(15)}, F_{\min}^{(20)} \right\}\), \(\mathcal{M}^{(t)} = \left\{ F_{0.5\lambda}^{(5)}, F_{0.5\lambda}^{(10)}, F_{0.5\lambda}^{(15)}, F_{0.5\lambda}^{(20)} \right\}\). The criterion is activated only after sufficient search history has been accumulated. In particular, it requires \(t \geq l, t > n\left(5+\frac{100}{\lambda}\right), \frac{\mathrm{FE}-\mathrm{FE}^{\mathrm{last}}}{\lambda} > \frac{T_{\mathrm{tstag}}}{2}, |\mathcal{L}^{(t)}|>100, |\mathcal{L}^{(t)}|>2l\), where $\mathrm{FE}^{\mathrm{last}}$ denotes the number of function evaluations when the best-so-far objective value was last updated, and \(l= \max\left\{ \frac{T_{\mathrm{tstag}}}{10}, \frac{|\mathcal{L}^{(t)}|}{10} \right\}\). Once these conditions are met, the criterion compares the historical and recent behavior of both the minimum and median offspring objective values, and stops when they have remained essentially unchanged for a sufficiently long period, indicating search stagnation.

\subsection{Benchmark stopping criteria for MOEAs}

\paragraph{MGBM}\cite{marti2016stopping} monitors the improvement between two consecutive populations using the mutual domination rate (MDR), 
$ \mathrm{MDR}\!\left(\mathcal{P}^{t},\mathcal{P}^{t-1}\right) = \frac{\left|\Delta\!\left(\mathcal{P}^{t-1},\mathcal{P}^{t}\right)\right|} {\left|\mathcal{P}^{t-1}\right|} - \frac{\left|\Delta\!\left(\mathcal{P}^{t},\mathcal{P}^{t-1}\right)\right|} {\left|\mathcal{P}^{t}\right|}$, \noindent where $\Delta(\mathcal{A},\mathcal{B})$ denotes the subset of solutions in $\mathcal{A}$ that are dominated by at least one solution in $\mathcal{B}$. A larger MDR indicates greater improvement from iteration $t-1$ to $t$. Since the observed MDR is assumed to be affected by Gaussian noise, MGBM applies a Kalman filter to sequentially estimate the underlying MDR. The search is terminated when this estimated improvement falls below a predefined threshold $\hat{I}_{\min}$. Thus, MGBM detects convergence by identifying when the estimated improvement between consecutive populations becomes sufficiently small.

\paragraph{Entropy-based Stopping Criterion (ESC)} \cite{saxena2015entropy} detects convergence by measuring the dissimilarity between two consecutive populations, $\mathcal{P}^{t-1}$ and $\mathcal{P}^{t}$, in the objective space. For an objective problem with $m$ objectives, the objective space is partitioned into $n_b^m$ grid cells. Let $\mathcal{C}_j$ denote the $j$th cell and $\mu$ the population size. The probability associated with cell $\mathcal{C}_j$ at iteration $t$ is estimated as $ p_j^{t} = \frac{n_j^{t}}{\mu}$, where $n_j^{t}$ is the number of objective vectors from $\mathcal{P}^{t}$ falling into $\mathcal{C}_j$. ESC then measures the dissimilarity between the distributions induced by $\mathcal{P}^{t-1}$ and $\mathcal{P}^{t}$. The search is terminated when this dissimilarity remains unchanged for a predefined number of iterations, $n_s$. Thus, ESC identifies convergence when the distribution of the population in the objective space becomes stable.

\paragraph{$\epsilon$-Progress-based Stopping Criterion ($\epsilon$SC)}\cite{hadka2013borg} monitors search progress using an $\epsilon$-box dominance archive. The objective space is partitioned into hyper-boxes of side length $\epsilon$, and solutions are compared according to the boxes they occupy. For example, given two objective vectors $\mathbf{u} = (u_1, \dots u_m)^\top$ and $\mathbf{v} = (v_1, \dots v_m)^\top$, $\mathbf{u}$ is said to $\epsilon$-box-dominate $\mathbf{v}$ if $\left\lfloor \frac{\mathbf{u}}{\epsilon} \right\rfloor \prec \left\lfloor \frac{\mathbf{v}}{\epsilon} \right\rfloor$, or $\left\lfloor \frac{\mathbf{u}}{\epsilon} \right\rfloor = \left\lfloor \frac{\mathbf{v}}{\epsilon} \right\rfloor$ and $\left\| \mathbf{u} - \left\lfloor \frac{\mathbf{u}}{\epsilon} \right\rfloor \right\| < \left\| \mathbf{v} - \left\lfloor \frac{\mathbf{v}}{\epsilon} \right\rfloor \right\|$ is satisfied. The archive retains representative non-dominated solutions from the occupied $\epsilon$-boxes, providing a coarse approximation of the Pareto front. Let $c_{\epsilon}$ denote the progress counter. It is increased whenever a newly generated solution is not dominated by the current archive and lies in a previously unoccupied $\epsilon$-box. The search is terminated if $c_{\epsilon}$ is not updated for $T_{\epsilon}$ consecutive iterations. Thus, $\epsilon$SC detects convergence when the search no longer discovers new non-dominated regions of the objective space.

\section{Additional Experimental Results}
\begin{figure}[!htb]
\centering
\includegraphics[width=1.0\textwidth]{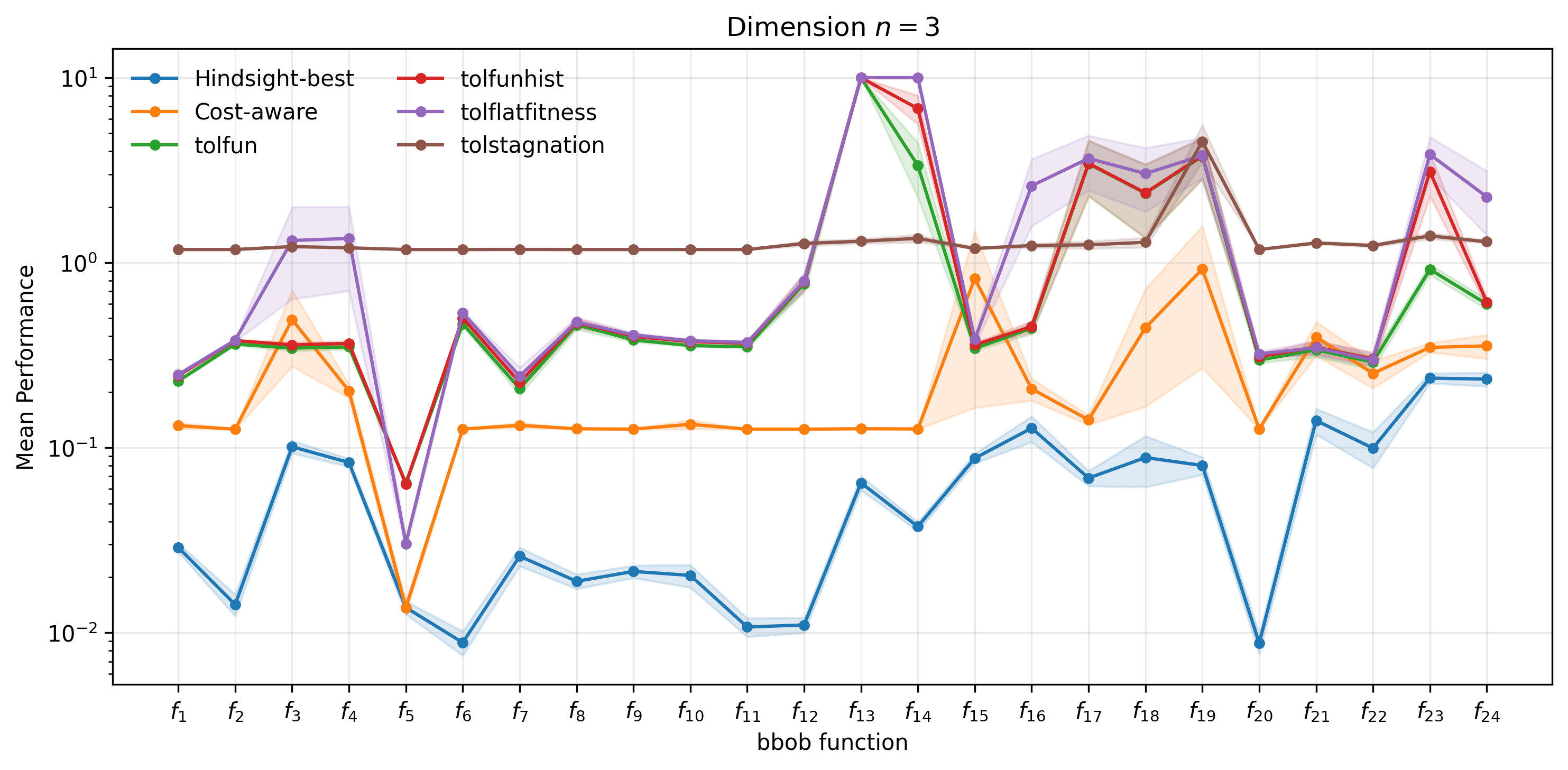}
\includegraphics[width=1.0\textwidth]{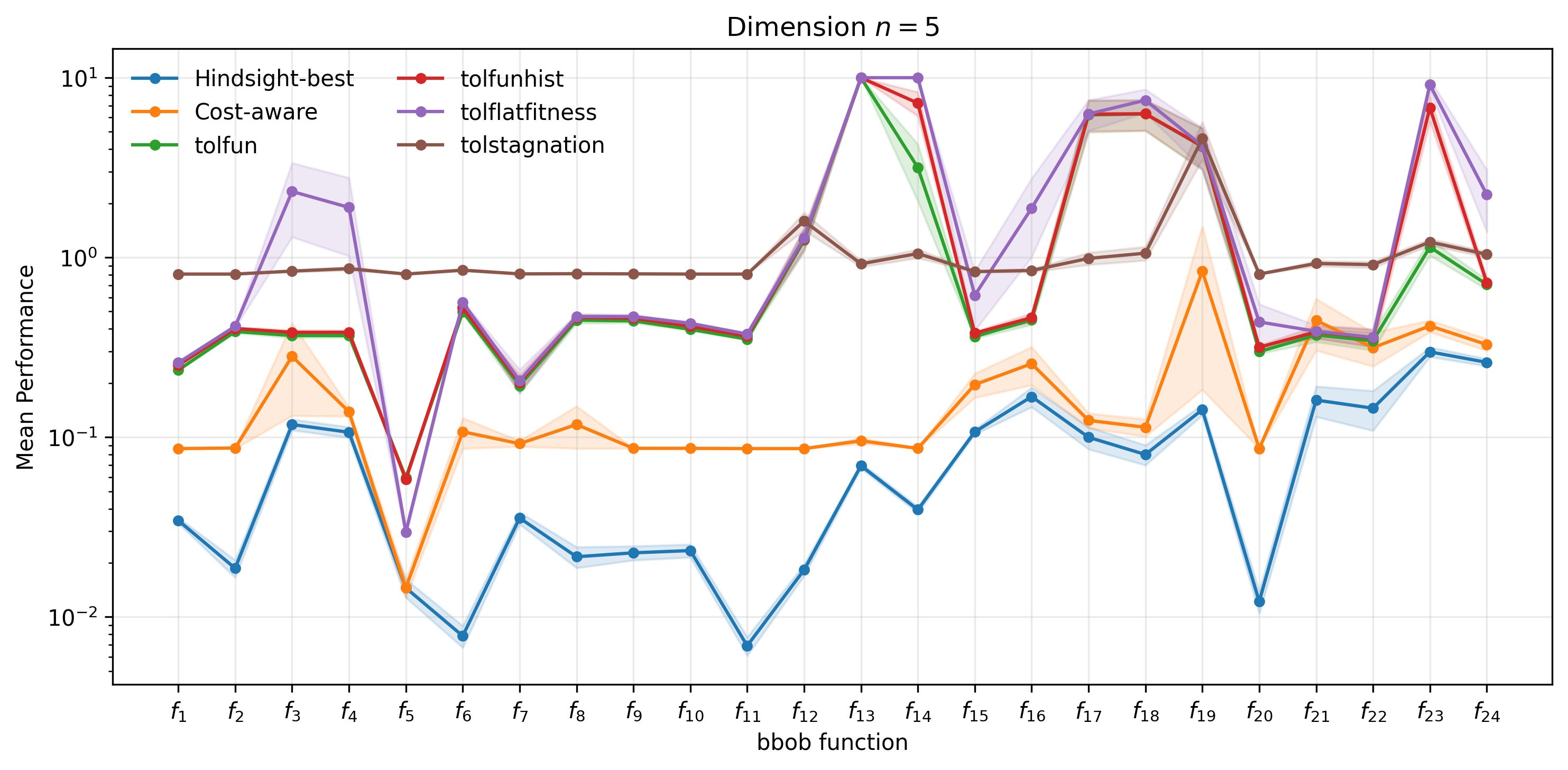}
\caption{SOEA mean normalized cost-aware performance of the seven stopping methods across the 24 \texttt{bbob} functions for dimensions $n=3$ and $n=5$. For each function and stopping method, the solid line shows the mean performance over 15 problem instances, and the shaded region represents $\pm$ one standard error. Lower performance indicates better stopping performance. } 
\label{fig:soea_performance_appendix}
\end{figure}

\begin{figure}[!htbp] 
\centering
\includegraphics[width=0.8\textwidth]{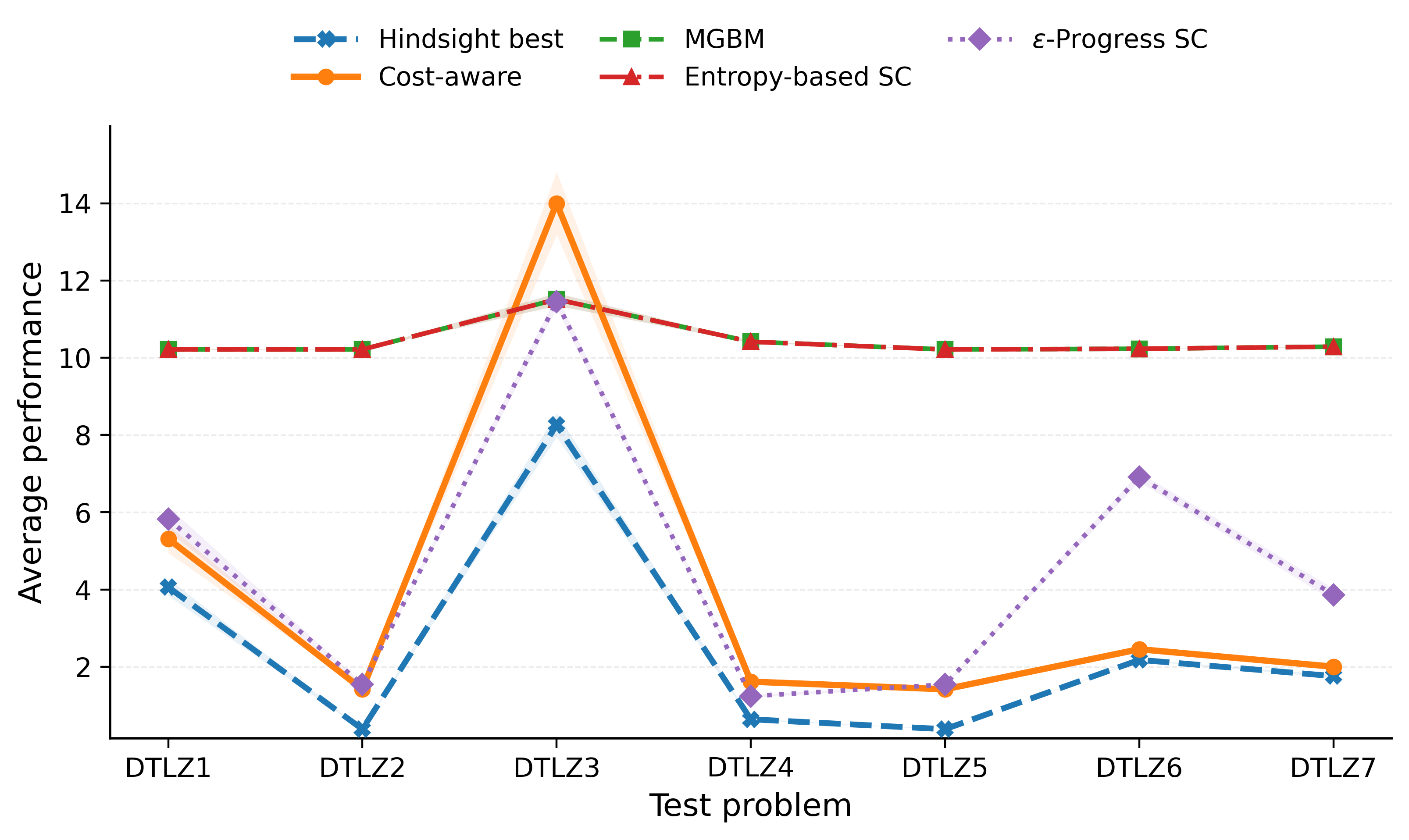}
\includegraphics[width=0.8\textwidth]{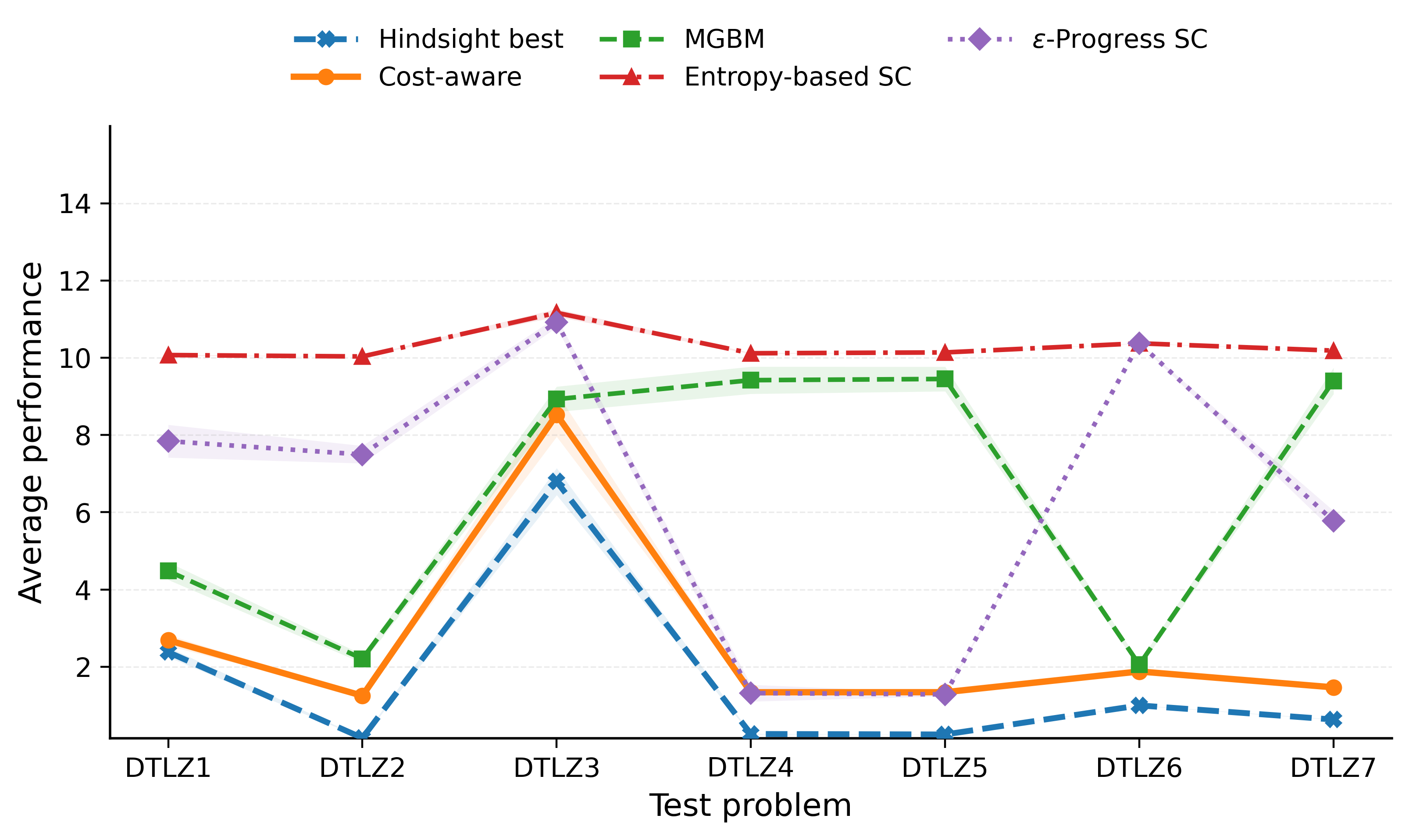}
\caption{MOEA mean normalized cost-aware performance of the stopping criteria across DTLZ1–DTLZ7 problems for 2(top) and 5(bottom) objectives. The line shows the mean performance over 31 replications, and the shaded region represents $\pm$ one standard error. Lower values indicate better stopping performance.} 
\label{fig:moea_performance_appendix}
\end{figure}

Here, we provide additional results for the experimental settings omitted from the main text for brevity. For SOEA, we report the normalized cost-aware performance across the 24 \texttt{bbob} functions for dimensions $n=3$ and $n=5$ in Figure~\ref{fig:soea_performance_appendix}. For MOEA, we report the corresponding cost-aware performance across DTLZ1-DTLZ7 for $m=2$ and $m=5$ in Figure~\ref{fig:moea_performance_appendix}. 

\section{Ablation Study of Elitist Prediction Condition}
Recall that the prediction model includes an elitist prediction constraint, 
\[ g_{\mathrm{comb}}(t+1\mid\boldsymbol{\xi}) \leq B_t, \]
as given in Constraint~\eqref{eq:curve_prediction_constraint}. This constraint ensures that the predicted next best-so-far objective value is no worse than the current best-so-far value. We conduct two ablation studies to assess the effect of this constraint on prediction accuracy and stopping performance.

\subsection{A prediction example without the elitist prediction constraint.} 
\begin{figure}[!htbp]
\centering
\includegraphics[width=0.8\textwidth]{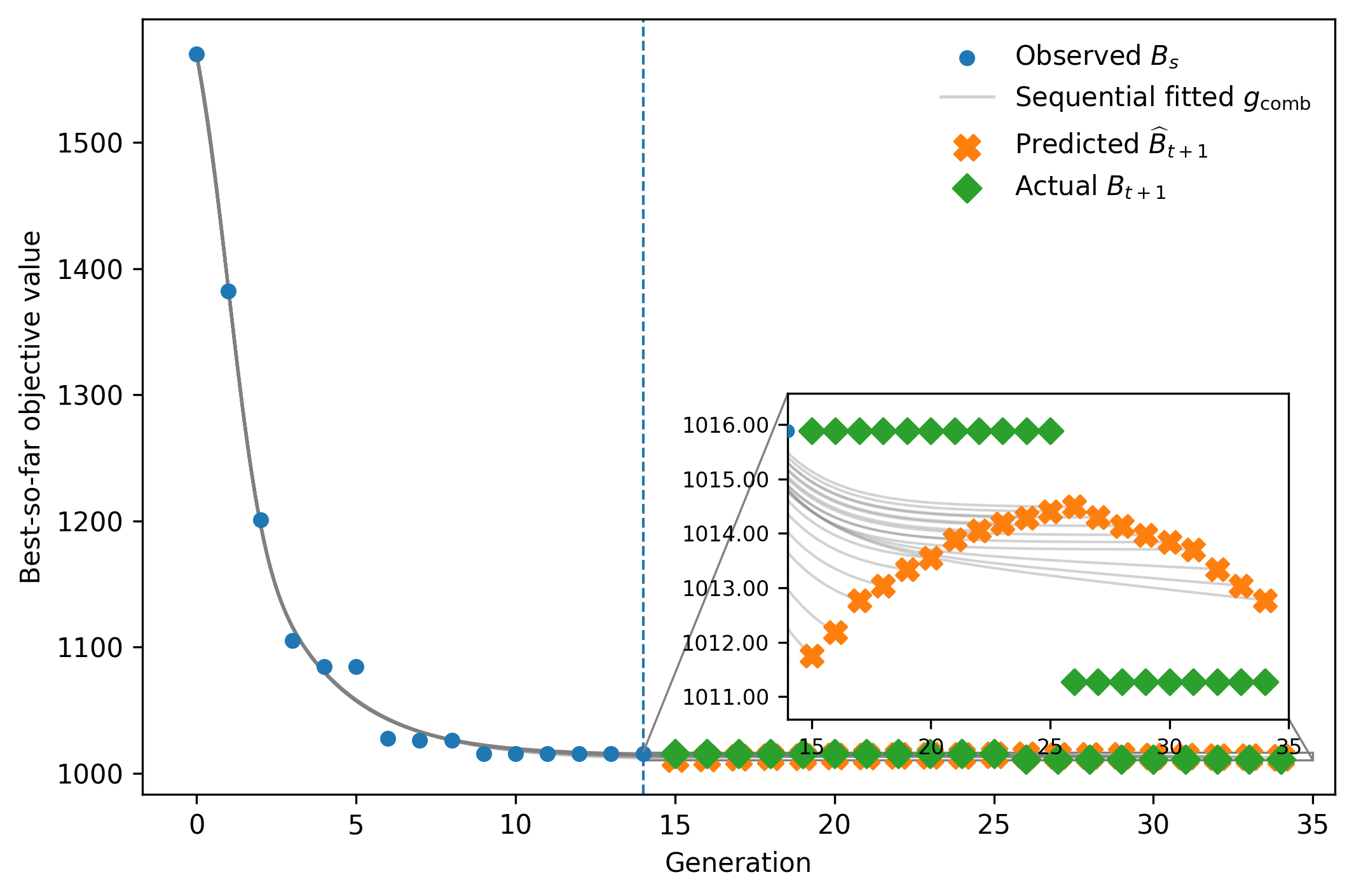}
\caption{A representative prediction example for CMA-ES on one instance of $f_{15}$ in the \texttt{bbob} suite with $n=3$, without the elitist prediction constraint in Eq.~\ref{eq:curve_prediction_constraint}. Prediction starts at generation 15 and continues for the next 20 generations. For the final fitted curve at generation 35, the model weights are $\omega_1=0.61$, $\omega_2=0.00$, $\omega_3=0.36$, and $\omega_4=0.03$.} 
\label{fig:regression_example_ablation}
\end{figure}

We perform a same illustrative prediction experiment from Section~\ref{sec:criterion} by removing the elitist prediction constraint from the learning-curve fitting procedure. The results are shown in Figure~\ref{fig:regression_example_ablation}. Compared with Figure~\ref{fig:regression_example} with the condition, it demonstrates that the constraint improves prediction consistency. Without the constraint, after the sharp improvement around generation 26, the predicted next best-so-far values become worse than the current best-so-far value, causing a clear prediction mismatch. In contrast, the constrained model respects the non-increasing best-so-far trajectory and adapts immediately to the new objective level.

\subsection{Stopping results without the elitist prediction constraint.}

To evaluate the effect of the elitist prediction constraint, we conduct ablation studies for both SOEA and MOEA. In each case, we remove the elitist prediction constraint from the learning-curve fitting procedure while keeping all other experimental settings unchanged, as described in Section~\ref{sec:setup}.

\subsubsection{SOEA.} We consider CMA-ES on the 24 \texttt{bbob} functions with dimension $n=2$, removing the constraint $\hat{L}_{t+1}\leq L_t$ while retaining the non-negativity constraint $\hat{L}_{t+1}\geq 0$. The results are reported in Table~\ref{tab:soea_n2_ablation}. Removing the constraint slightly changes the mean performance of the cost-aware criterion from $0.22 \pm 0.03$ (as shown in Table~\ref{tab:soea_performance_summary}) to $0.21 \pm 0.01$ and reduces the mean stopping FEs from $389.37 \pm 55.09$ (as shown in Table~\ref{tab:soea_stopping_summary}) to $366.20 \pm 24.50$.

\begin{table}[!htbp]
\centering
\caption{Mean performance and stopping function evaluations (FEs)
of the stopping criteria for CMA-ES on the 24 \texttt{bbob} functions with dimension $n=2$, without the elitist prediction constraint. Results are reported as mean $\pm$ one standard error.
Lower values indicate better performance and fewer evaluations,
respectively.}
\label{tab:soea_n2_ablation}
\begin{tabular}{l C{4cm} C{5cm}}
\toprule
\textbf{Stopping criterion} & \textbf{Mean performance} & \textbf{Mean stopping FEs} \\
\midrule
\texttt{tolfun}         & $0.69 \pm 0.08$ & $1331.08 \pm 161.62$ \\
\texttt{tolfunhist}     & $0.96 \pm 0.11$ & $1859.60 \pm 216.48$ \\
\texttt{tolflatfitness} & $1.17 \pm 0.13$ & $2282.35 \pm 254.78$ \\
\texttt{tolstagnation}  & $1.57 \pm 0.01$ & $3081.57 \pm 17.77$  \\
Cost-aware              & $\textbf{0.21}\pm 0.01$ & $\textbf{366.20} \pm 24.50$ \\
Hindsight-best*         & $0.06 \pm 0.01$ & $50.40 \pm 2.43$ \\
\bottomrule
\end{tabular}
\end{table}

\begin{table}[!htb]
\centering
\caption{Mean performance and stopping function evaluations (FEs) of the stopping criteria for MOEA/D with $m=3$ objectives.
Results are reported as mean $\pm$ one standard error. Lower values indicate better performance and fewer evaluations, respectively.}
\label{tab:moea_m3_ablation}
\begin{tabular}{l C{4cm} C{5cm}}
\toprule
\textbf{Stopping criterion} & \textbf{Mean performance} & \textbf{Mean stopping FEs} \\
\midrule
MGBM & $9.65 \pm 0.14$ & $9107.83 \pm 106.83$ \\
Entropy-based SC & $10.54 \pm 0.07$ & $10000.00 \pm 0.00$ \\
$\epsilon$-Progress SC & $6.18 \pm 0.29$ & $5556.45 \pm 249.24$ \\
Cost-aware & $\textbf{3.58} \pm 0.30$ & $\textbf{1240.09} \pm 14.12$ \\
Hindsight best & $2.28 \pm 0.21$ & $874.88 \pm 111.48$ \\
\bottomrule
\end{tabular}
\end{table}

\subsubsection{MOEA.} We conduct the same ablation study for MOEA/D with $m=3$ objectives by removing the constraint $\hat{R}_{t+1}\leq R_t$. As shown in Table~\ref{tab:moea_m3_ablation}, the mean performance of the cost-aware criterion slightly worsens from $3.54 \pm 0.30$ (as shown in Table~\ref{tab:moea_performance_summary}) to $3.58 \pm 0.30$, while the mean stopping FEs decrease from $1289.63 \pm 29.83$ (as shown in Table~\ref{tab:moea_stopping_summary}) to $1240.09 \pm 14.12$. 
Unlike SOEA, the standard error of performance remains unchanged in MOEA, while that of stopping FEs decreases from $29.83$ to $14.12$, suggesting that removing the constraint reduces variability in stopping FEs without noticeably affecting performance variability.

Removing the elitist prediction constraint leads to earlier stopping in both SOEA and MOEA experiments. It marginally improves performance in SOEA but slightly worsens performance in MOEA, none of the differences is statistically significant.
Overall, while the elitist prediction constraint seems logical, these preliminary results suggest that its benefit may be small and depend on the problem characteristics such as the trade-off between solution improvement and evaluation costs.

\end{document}